\documentclass[lettersize,journal]{IEEEtran}
\usepackage{amsmath,amsfonts}
\usepackage{algorithmic}
\usepackage{algorithm}
\usepackage{array}
\usepackage[caption=false,font=normalsize,labelfont=sf,textfont=sf]{subfig}
\usepackage{textcomp}
\usepackage{stfloats}
\usepackage{url}
\usepackage{verbatim}
\usepackage{graphicx}
\usepackage{cite}
\usepackage{bm}
\usepackage{amsmath}
\usepackage{pifont}
\usepackage{amsmath}
\usepackage{graphicx}
\usepackage{multirow}
\usepackage{makecell}
\usepackage{xcolor}
\usepackage{booktabs}
\usepackage{hyperref}
\usepackage{stfloats}
\usepackage{caption}

\newcommand{\etal}{\textit{et al}.}

\newcommand{\ie}{\textit{i}.\textit{e}.}
\newcommand{\eg}{\textit{e}.\textit{g}.}

\begin{document}

\title{Self-supervised Learning of Event-guided Video Frame Interpolation for Rolling Shutter Frames}

\author{Yunfan Lu$^*$, Guoqiang Liang$^*$, Yiran Shen and Lin Wang$\dagger$
\thanks{Yunfan Lu and Guoqiang Liang are co-first authors and with AI Thrust, HKUST(GZ). E-mail: \{ylu066,gliang041\}@connect.hkust-gz.edu.cn}
\thanks{Yiran Shen is with school of software Shandong University. E-mail: yiran.shen@sdu.edu.cn.}
\thanks{Lin Wang is the corresponding author and with School of Electrical and Electronic Engineering, Nanyang Technological University, Singapore. E-mail: linwang@ntu.edu.sg.}
}

\markboth{Journal of \LaTeX\ Class Files,~Vol.~14, No.~8, August~2021}%
{Shell \MakeLowercase{\textit{et al.}}: A Sample Article Using IEEEtran.cls for IEEE Journals}

\IEEEpubid{}

\twocolumn[{
\renewcommand\twocolumn[1][]{#1}%
\maketitle
\vspace{-20pt}
\begin{center}
\centering
\includegraphics[width=0.8\linewidth]{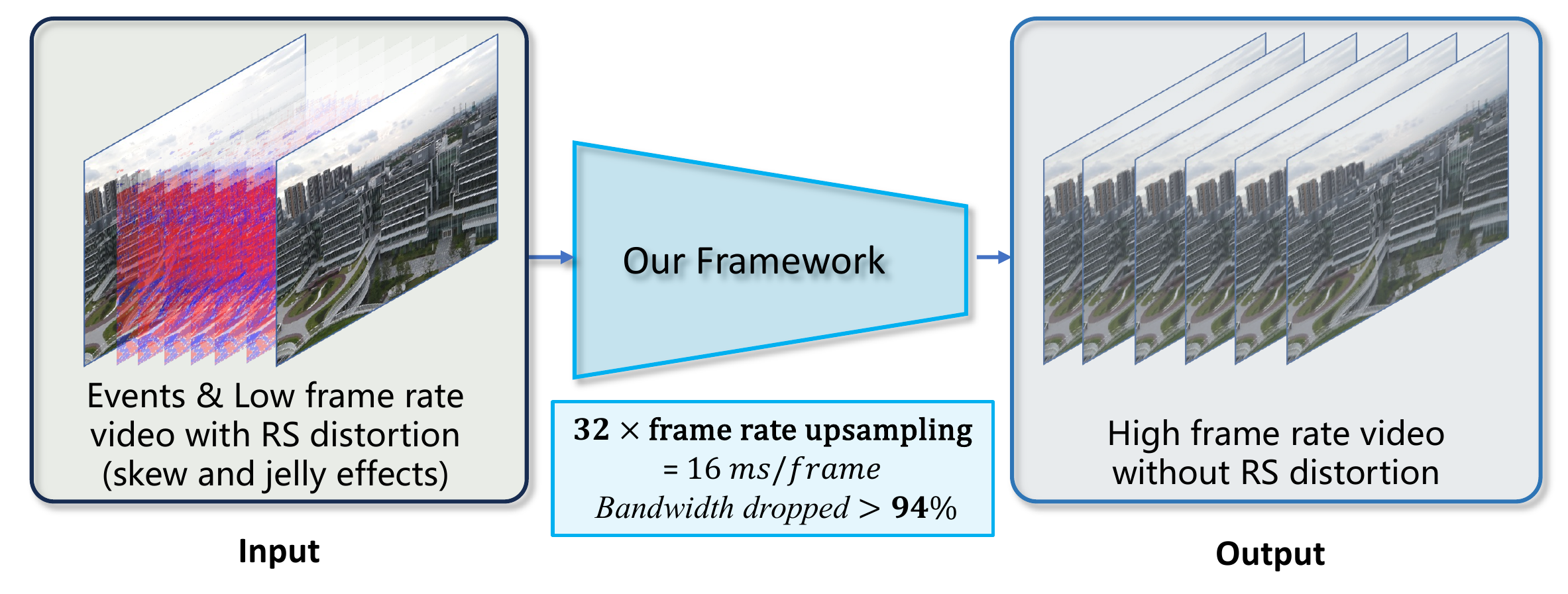}
\end{center}
\captionof{figure}{\small In this paper,  we introduce a novel approach that allows for streaming high frame rate global shutter (GS) videos without distortion (\eg, skew and jelly effects) from a {multi-sensor-equipped system(\ie, hybird camera with rolling shutter (RS) RGB and event sensor~\cite{yunfan2024rgb,zhou2025binarized})}.
Due to the lack of labeled datasets for supervised training, our framework is self-supervised, buttressed by the GS-to-RS mutual reconstruction strategy.
Both object and subjective analysis demonstrate that our approach achieves over $\bm{94\%}$ bandwidth reduction compared with the high frame rate video and compared with the event-based methods~\eg, TimeLens~\cite{tulyakov2021time}+EvUnroll~\cite{zhou2022evunroll}, where each frame is in an average of just $\bm{16~ms/frame}$ under frame interpolation with $32 \times$ frame rate upsampling.
Bandwidth efficiency and low time consumption make our approach potential well-suited for many applications.}
\label{fig:05-differ}
\vspace{10pt}
}]

\begin{abstract}
Most consumer cameras use rolling shutter (RS) exposure, the captured videos often suffer from distortions (\eg, skew and jelly effect).
Also, these videos are impeded by the limited bandwidth and frame rate, which inevitably affect the video streaming experience.
In this paper, we excavate the potential of event cameras as they enjoy high temporal resolution.
Accordingly, we propose a framework to recover the global shutter (GS) high frame rate (\ie, slow motion) video without RS distortion from an RS camera and event camera.
One challenge is the lack of real-world datasets for supervised training.
Therefore, we explore self-supervised learning with the key idea of estimating the displacement field—a non-linear and dense 3D spatiotemporal representation of all pixels during the exposure time.
This allows for a mutual reconstruction between RS and GS frames and facilitates slow-motion video recovery.
We then combine the input RS frames with the DF to map them to the GS frames (\textit{RS-to-GS}).
Given the under-constrained nature of this mapping, we integrate it with the inverse mapping (\textit{GS-to-RS}) and RS frame warping (\textit{RS-to-RS}) for self-supervision.
We evaluate our framework via objective analysis (\ie, quantitative and qualitative comparisons on four datasets) and subjective studies (\ie, user study).
The results show that our framework can recover slow-motion videos without distortion, with much lower bandwidth ($94\%$ drop) and higher inference speed ($16ms/frame$) under $32 \times$ frame interpolation.
The dataset and source code are publicly available at:~{\url{https://github.com/yunfanLu/Self-EvRSVFI}}.
\footnotetext[1]{Yunfan Lu and Guoqiang Liang are co-first authors, with AI Thrust, HKUST(GZ). Email: \{ylu066, gliang041\}@connect.hkust-gz.edu.cn}
\footnotetext[2]{Yiran Shen is with the School of Software, Shandong University. Email: yiran.shen@sdu.edu.cn}
\footnotetext[3]{Lin Wang (corresponding author) is with the School of Electrical and Electronic Engineering, Nanyang Technological University, Singapore. Email: linwang@ntu.edu.sg}
\end{abstract}


\section{Introduction}

The burgeoning interest in virtual reality (VR) has recently been extended to most consumer cameras, \eg unmanned aerial vehicles (UAVs), offering users an unprecedented and immersive flight experience~\cite{bacco2020monitoring,zhang2019augmented,ponnusamy2021precision,li2021log}.
High-frame-rate (\ie, slow-motion) videos free from distortions are pivotal for achieving this level of immersion~\cite{lu2023bh,omori2018120,tan2018360,kamarainen2018cloudvr,vieri201818}.
However, attaining such an ideal video is constrained by inherent hardware limitations~\cite{oagaz2021performance,cross2022using,danyluk2020touch}.
Specifically, the rolling shutter (RS) sensors,  which are commonly used in most consumer cameras, introduce distortions, \eg, skew and the jelly effect, particularly during rapid movements~\cite{fan2022context,chun2008suppressing,hedborg2012rolling,schubert2019rolling}.
While the global shutter (GS) sensors circumvent RS distortion, they come at a cost that is more than ten times higher than RS sensors and also consume significant power~\cite{tan2018360}.
Additionally, constraints on bandwidth~\cite{ding2021uav,luo2023masked360} and computational speed~\cite{leng2019energy,omori2018120,luo2023masked360,friess2020foveated,betancourt2022exocentric} also should be considered for VR video streaming.
Against this backdrop, we aim to enhance video frame rate and correct RS distortion by introducing a framework characterized by fast inference speed and low bandwidth.

In light of these hardware constraints, there has been a growing demand for software-based solutions to achieve high-quality, high-frame-rate videos.
This has propelled deep learning-based video frame interpolation (VFI) methods to the forefront of research~\cite{jiang2018super,wu2022video}.
However, it's worth noting that despite the success of these VFI methods~\cite{meyer2018phasenet,jiang2018super,niklaus2018context,xu2019quadratic,niklaus2020softmax,chi2020all,park2020bmbc,paliwal2020deep}, they predominantly operate under the GS mechanism, which doesn't directly address the RS distortions commonly found in UAV-captured videos, especially the high-speed or dynamic motion scenes~\cite{tulyakov2021time,lu2024hr}.
To tackle this problem, learning-based RS correction methods~\cite{Liu2020DeepSU,rengarajan2016bows,zhuang2019learning} have been proposed to obtain GS frames by removing the RS effect based on the assumption of linear motion.
However, they can not synthesize nonexistent in-between GS frames.
Therefore, it is desirable to generate in-between GS frames from two consecutive RS frames, as it can benefit both VFI and RS correction.
Accordingly, some learning-based methods have been proposed. RSSR~\cite{fan2021inverting} and CVR~\cite{fan2022context} are two representative methods that estimate the linear motion to recover faithful in-between GS frames from two consecutive RS frames.
However, they can not handle non-linear motion, which often occurs in dynamic scenes with fast motion.

Event cameras are bio-inspired sensors that can asynchronously detect per-pixel intensity changes and generate event streams with high temporal resolution\textemdash 1$us$ and high dynamic range compared with the frame-based cameras~\textemdash 140$dB$ vs. 60$dB$ with low power consumption~\textemdash $10 mW$~\cite{hu2021event,scheerlinck2019ced,gallego2020event,zheng2023deep,lu2023learning}.
This has inspired some research endeavors~\cite{tulyakov2021time,lu2025events,he2022timereplayer,paikin2021efi,yu2021training,wu2022video,feng2022real}, exploring event cameras as the guidance to compensate for motion information loss for VFI in the dynamic scenes with fast motion.
However, these methods only serve the goal of GS frame interpolation via supervised learning.
Moreover, they necessitate paired GS and RS frame data, which are typically acquired via simulation on high-speed videos~\cite{fan2022context,fan2021inverting,li2021deep}.
This not only incurs substantial costs for components of expensive optical equipment~\cite{tulyakov2021time,tulyakov2022time} but also restricts the practical usage of these techniques to UAV-VR video streaming.

\begin{figure*}[t!]
\centering
\includegraphics[width=0.89\linewidth]{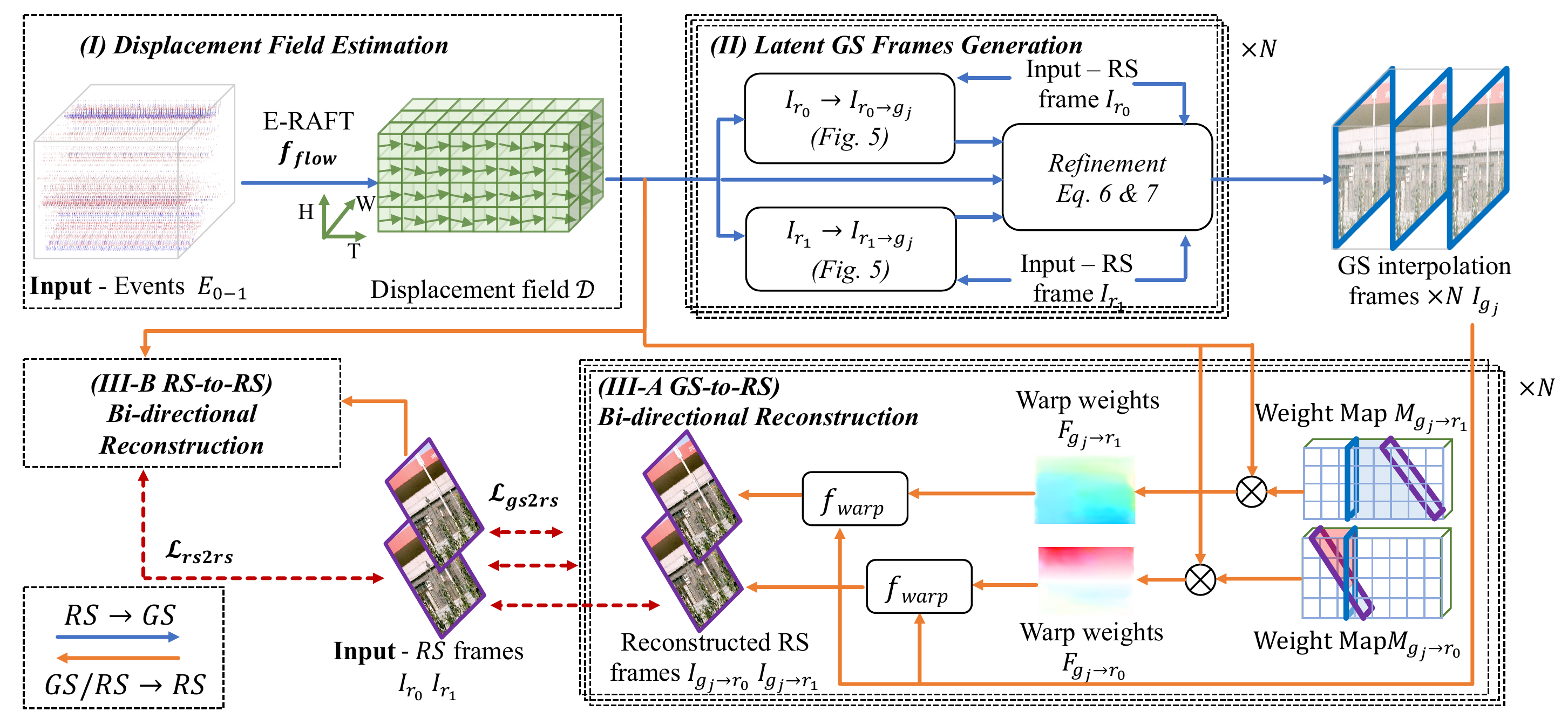}
\caption{{Overview of our proposed framework, which consists of three parts, (I) the displacement field estimation module, (II) latent GS frames generation module, and (III) the reciprocal reconstruction module. {Additional context and explanations are provided in the main text.}}}
\label{fig:overall_framework}
\centering
\end{figure*}

In this paper, we make the first attempt to {leverage the high temporal resolution of event cameras to guide the recovery of in-between GS frames from two consecutive RS frames based on a multi-sensor-equipped UAV}.
However, tackling this novel problem is non-trivial because \textbf{1)} there are no GS-events-RS triplet datasets for VFI, and \textbf{2)} RS frames are suspectable to edge distortion and region occlusion in dynamic scenes with fast motion.
To this end, we propose a novel
{self-supervised learning} (SSL) framework that leverages events to guide RS correction and VFI in a unified framework.
Overall, our method enables the recovery of GS frames with {any arbitrary frame rate, \eg, 32$\times$,} from two consecutive RS frames, guided by events, as depicted in Fig.~\ref{fig:05-differ}.
The proposed SSL framework is shown in Fig.~\ref{fig:overall_framework}.
The key idea of our method is to 1) estimate the 3D {displacement field} (DF), which includes dense spatiotemporal non-linear motion information of all pixels during the exposure time, and 2) combine the RS frames and DF for the reciprocal reconstruction (or mapping) to impose self-supervision.

Specifically, we first propose the displacement field estimation module to estimate the spatiotemporal motion information from events directly (Sec.~\ref{sec:displacement_field_estimation}).
We split events into moments, each of which includes a fixed amount of voxel grid~\cite{stanescu2022model,rebecq2018emvs,gehrig2019end}.
We estimate the optical flow~\cite{gehrig2021raft} between consecutive event moments. This way, we can obtain the non-linear 3D motion information\textemdash DF, during the exposure time for GS frames.
Benefiting from the high temporal resolution of events, 3D DF contains dense motion information for RS correction and GS frames interpolation in one step.
Based on DF, we propose a latent GS frames generation module to learn the RS-to-GS mapping for GS frame interpolation (Sec.~\ref{sec:latent_gs_frame_generation}).
We generate a series of GS frames at arbitrary frame rate in exposure time from RS frames and 3D DF.
As the ground truth GS frames are not available, the mapping is highly under-constrained. Thus, we couple it with a
reciprocal reconstruction module to 1) reconstruct RS frames based on the generated GS frames and fully exploit the constraints inherent in RS frames, and 2) achieve RS-to-RS warping based on the DF for self-supervision (Sec.~\ref{sec:self-supervision}).

Due to the lack of RS-events-GS triplet datasets, we generate two simulated datasets for qualitative and quantitative evaluation.
We propose the \textbf{{first}} real-world dataset with RS frames and aligned events for training and evaluation of our framework. Also, we collect a dataset using a UAV (equipped with RGB and event cameras) to validate our framework's effectiveness.
We conduct objective and subjective evaluations for our framework, focusing on key metrics such as inference speed, bandwidth efficiency, generalization performance.
The evaluation results show that our method yields
\textbf{1}) {comparable} performance with supervised frame-based RS-to-GS VFI method~\cite{fan2022context} and
\textbf{2}) {better} performance than the supervised event-guided RS correction~\cite{zhou2022evunroll} + unsupervised event-guided VFI method\textemdash TimeReplayer~\cite{he2022timereplayer}.
Note that with RS frames and events as inputs, our method achieves much higher VFI performance than that of TimeLens (only for GS frames)~\cite{tulyakov2021time}.
In practice, the results underscore the remarkable prowess of our method in restoring slow-motion videos to their pristine quality, achieving this feat with a remarkable $94\%$ reduction in bandwidth usage and $16~ms$ per frame inference speed in the demanding scenario of $32 \times$ frame interpolation for UAV-VR application purpose.

In summary, the contributions of this paper are four-fold: (\textbf{I}) we propose the {first} self-supervised approach to recover GS slow-motion video from a multi-sensor-equipped UAV. (\textbf{II}) We introduce a DF estimation module and a reciprocal reconstruction module to impose self-supervision. \textbf{(III)} We propose the first real-world RS-event paired dataset for the training and evaluation.
\textbf{(IV)} We conduct analysis of our approach, showcasing the benefits of our method with respect to bandwidth efficiency and inference speed.
Collectively, our approach serves as a technical exploration, opening up possibilities for improving the UAV-VR video streaming experience by correcting RS distortion and enhancing frame rate.

\section{Related Works}

The growing fascination with VR has expanded its horizons into the realm of UAVs, promising users an unparalleled and immersive flying experience~\cite{bacco2020monitoring,zhang2019augmented,ponnusamy2021precision,wu2021spinpong}.
Achieving a heightened sense of immersion in this context heavily relies on the availability of high-frame-rate (\ie, slow-motion) videos devoid of distortions ~\cite{lu2023bh,omori2018120,tan2018360,kamarainen2018cloudvr,vieri201818}.
Our primary focus is on harnessing events to guided RS frame correction and interpolation, leading us to categorize recent technical research into the following three facets:

\noindent\textbf{Event-guided VFI:}
Video frame interpolation (VFI) is a fundamental task converting low frame rate videos to high frame rate videos in video enhancement~\cite{meyer2018phasenet,dong2022video,parihar2022comprehensive}.
During generating intermediate frames between consecutive frames, frame-based methods~\cite{meyer2018phasenet,jiang2018super,niklaus2018context,xu2019quadratic,niklaus2020softmax,chi2020all,park2020bmbc,paliwal2020deep} predict motion explicitly or implicitly.
However, these methods degrade greatly in scenes of non-linear motion since accurately modeling motion from the sparse set of frames is ill-posed~\cite{yu2021training}.
Unlike the RGB camera, the event camera enjoys many advantages, such as high temporal resolution, which captures the brightness changes in a time interval~\cite{tulyakov2021time}.
Previous works have demonstrated its potential for VFI, and they can be roughly categorized into two types: supervised and unsupervised methods.
The supervised methods can also be divided into two parts: synthesis-based and wrapping-based methods~\cite{paikin2021efi,wu2022video}.
Time Lens~\cite{tulyakov2021time} and Time Lens++~\cite{tulyakov2022time} combine synthesis methods with warping-based methods to boost the VFI performance.
TimeReplayer~\cite{he2022timereplayer} is an unsupervised approach that applies a loss between the input frames and reconstructed input frames\textemdash warped from interpolated frames.
{All these methods are designed for the GS frames without considering the distortion caused by RS.
However, most commercial cameras record frames with the RS mechanism, thus impeding their applications in real-world scenarios.}

\noindent\textbf{RS correction:}
Recently, some learning-based methods have been proposed to achieve the RS correction~\cite{dibene2022prepare,Liu2020DeepSU,Fan2021InvertingAR,Fan2021SUNetSU,cao2022learning,fan2022context}.
As the motion information between frames is unknown, these methods rely on the prior assumption of linear motion to predict the intermediate GS frames, which degrade greatly in scenes of non-linear motion.
Zhou \etal~\cite{zhou2022evunroll} explored the spatiotemporal information of event cameras to boost the performance of RS correction with the non-linear motion for the target time.
{However, they focus on RS correction and deblurring without considering RS-to-GS frame interpolation.}

\noindent\textbf{VFI with RS frames:}
RSSR~\cite{fan2021inverting} proposes the first work to recover a random frame rate GS frame from RS frames, while the results suffer from unwanted holes and black edges, caused by occlusion between the RS and GS frames.
CVR~\cite{fan2022context} then proposes a context-aware GS frame interpolation framework to alleviate the occlusion problem and reduce artifacts.
However, these methods are frame-based and only focus on linear motion. It is demanding to consider non-linear motion for real-world applications.
In addition, these methods need the paired RS-GS dataset for training, which is difficult to collect.
Therefore, they only conduct experiments on the synthetic dataset.
{We make the first attempt to leverage events to guide RS correction and VFI in one step.
Accordingly, we propose a self-supervised approach that generates the arbitrary frame rate GS frames between consecutive RS frames.}

\begin{table}[t]
\caption{Definitions of Mathematical Symbols}
\centering
\renewcommand{\arraystretch}{1}
\begin{tabular}[width=1\linewidth]{l|l}
\hline
\textbf{Symbol} & \textbf{Definition} \\
\hline
$\bm{L}$ & Ideal Continuous Global Shutter (GS) Video Intensity \\
$I_r$ & Rolling Shutter (RS) Frame \\
$t^{r}_{s}$, $t^{r}_e$ & RS Frame Exposure Start/End Times \\
$\Delta t$ & RS Line Exposure Interval \\
$H$ & Video Frame Height \\
$t_h$ & RS Row $h$ Exposure Time \\
$E$ & Event Stream \\
$e$ & Single Event $(t, x, y, p)$ \\
$C$ & Event Triggering Threshold (Brightness) \\
$\bm{L}(t, x, y)$ & Brightness at Pixel $(x, y)$ at Time $t$ \\
$\Delta b$ & Brightness Change \\
$\Phi$ & Event Triggering Function \\
$p_e$ & Brightness Change Polarity (+/-) \\
$tr(t)$ & Point Trajectory $(x, y)$ during Exposure \\
$\Delta p_{0\to 1}$ & Displacement from Time $t_0$ to $t_1$ \\
$f_{flow}$ & Event-based Optical Flow Estimation Function \\
$\mathbf{D}$ & Displacement Field \\
$\mathbf{L}_{field}$ & Displacement Field Loss \\
$\nabla$ & Directional Gradient \\
$\mathbf{M}$ & Weight Map for RS/GS Transformation \\
$f_{warp}$ & Warping Function \\
$\mathbf{L}_{gs2rs}$ & GS-to-RS Self-Supervision Loss \\
$\mathbf{L}_{rs2rs}$ & RS-to-RS Self-Supervision Loss \\
$\mathbf{L}$ & Total Loss (Weighted Sum) \\
\hline
\end{tabular}
\label{tab:symbols}
\end{table}

\section{Preliminaries}
\label{subsec:foundation_of_rs_gs_events}
Rolling Shutter is an exposure mechanism widely deployed in commercial cameras, which determines the VFI problem definition.
{For clarity, the mathematical symbols used in this paper are defined in Tab.~\ref{tab:symbols}}.
First, we denote $\bm{L}$ as the intensity of an ideal video sequence in the period of $[t_s, t_e]$ containing continuous GS frames.
A RS frame $I_{r}$ can be regarded as a special composition of a series of GS frames. We denote the $t^{r}_{s}$ and $t^{r}_e$ as the start and end of the exposure time of RS frame $I_{r}$, respectively.
The interval of exposure time between every two adjacent lines is $\Delta t=(t^{r}_s - t^{r}_e)/(H-1)$, where $H$ is the height of video frames.
Therefore, the exposure time of row $h$ of the RS frame can be recorded as $t_h=t^{r}_{s}+h\times \Delta t$.
Then the raw $h$ of RS frame $I_{rs}$ can be formulated as Eq.~\ref{eq:rs_to_gs_v}, where $I_{t_h}[h]$ refers to the row $h$ of the GS frame captured at $t_h$.
{\begin{equation}
    I_{rs}[h] = I_{t_h}[h].
    \label{eq:rs_to_gs_v}
\end{equation}}

For an event stream $E$, which is a set of event $e=\{(t,x,y,p)\}$, each event is triggered and recorded when the brightness change exceeds a certain threshold $C$ at pixel $(x,y)$.
Denote the time interval as $\Delta t_e$, which is quite a short period, and the brightness at position $(x,y)$ as $\bm{L}(t,x,y)$, where $x\in [0,H]$ and $y\in [0,W]$.  The brightness change can be calculated as $\Delta b = log(\bm{L}(t_e,x_e,y_e)) - log(\bm{L}(t_e - \Delta t_e,x_e,y_e))$. Hence, the event at $t$ can be formulated as $p_e = \Phi(\Delta b, C)$, where $\Phi$ is the event triggering function. Event is recorded when $|\Delta b| > C$, and $p_e \in \{1,-1\}$ indicates the increase or decrease.

\begin{figure}[t!]
\centering
\caption{
{Illustration of the input RS frames (\textcolor{purple}{purple}) and events (a) and output GS frames after interpolation (\textcolor{blue}{blue}) for $4\times$ VFI (b).}}
\includegraphics[width=\linewidth]{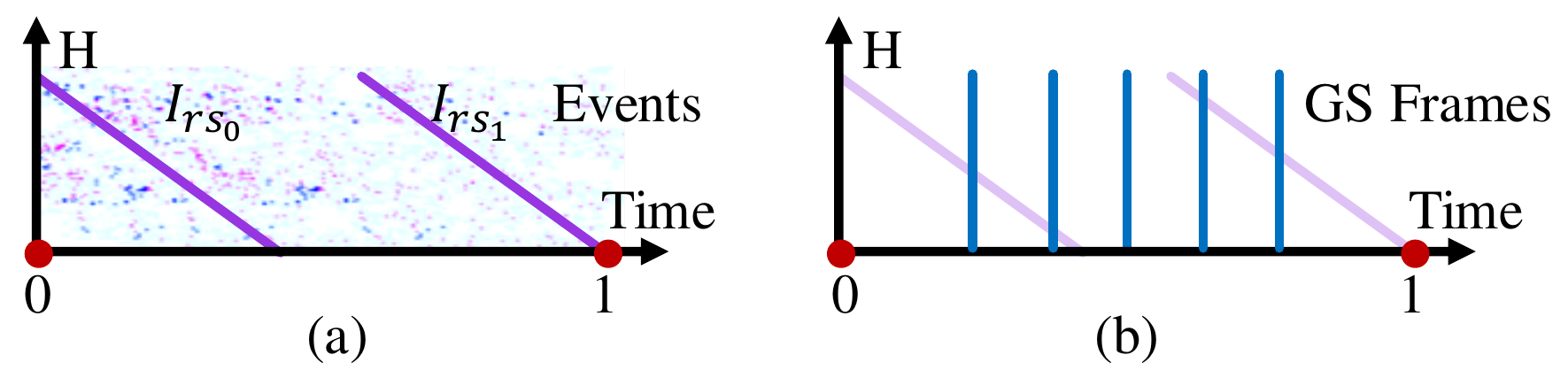}
\label{fig:input-output}
\end{figure}

\section{Proposed Framework}
\noindent \textbf{Overview:}
The overall framework of the proposed approach is depicted in Fig.~\ref{fig:overall_framework}, which can be divided into three parts:
(I) Displacement Field Estimation,
(II) Latent GS frames Generation, and
(III) Reciprocal Reconstruction.
As the oversimplified linear motion model fails in complex non-linear motions, we first introduce DF which contains non-linear motion during the exposure time and bridge the gap between RS frames and GS frames (Sec.~\ref{sec:displacement_field_estimation}), followed by describing how to generate latent GS frames by RS frames and DF (Sec.~\ref{sec:latent_gs_frame_generation}).
Since the mapping from RS to GS is highly under-constrained, we couple it with the inverse mapping (GS-to-RS) and RS frame warping (RS-to-RS) for self-supervision (Sec.~\ref{sec:self-supervision}).
The inputs of our framework are two consecutive RS frames($I_{r_0}$, $I_{r_1}$) and their corresponding events, while the outputs are continuous latent global shutter (GS) frames at arbitrary frame rate, as shown in Fig.~\ref{fig:input-output}.

\subsection{Displacement Field Estimation (DFE)}
\label{sec:displacement_field_estimation}

{This module aims to model non-linear motion information from events and lays a foundation for reciprocal reconstruction between GS and RS frames.}
Assume that the camera captures a 3D point in complex non-linear motion, and the trajectory of this point in the exposure time is denoted as $tr(t)=(x,y)$, where $t$ is the timestamp and $(x,y)$ is the pixel location.
Due to the non-linear character of $tr$, it can not be accurately expressed using previous frame-based VFI methods based on the assumption of linear motion~\cite{fan2021inverting,fan2022context}.
To this end, we decompose the whole trajectory into several pieces and use a piece-wise linear function to fit it.
Given a very short time period $[t_0,t_1]$, the pixel location $p_{t0}$ and $p_{t1}$ at these two timestamps are $tr(t_0)=(x_0,y_0)$ and $tr(t_1)=(x_1,y_1)$, respectively.
We use linear approximation to fit the motion during $t_0 - t_1$, then we can obtain $\Delta p_{0\to 1}= tr({t_1})-tr(t_0) = (x_1-x_0,y_1-y_0)$, which is the displacement from time $t_0$ to time $t_1$.
$\Delta p_{0\to 1}$ can be easily approximated by the estimation of the optical flow.
However, how to estimate the optical flow in a small time period $[t_0,t_1]$ is challenging.

To address this issue, we leverage the high-temporal resolution of events to estimate the optical flow within $[t_0,t_1]$.
We divide the events into $T+1$ time bins, and each time bin is further divided into $N$ voxel grids~\cite{rebecq2018emvs,wang2020eventsr}.
Based on this, we can get the event representation with a dimension of $(T+1)\times N \times H \times W$.
We denote the event-based optical flow estimation function as $f_{flow}$, and the dimension of the estimated optical flow set is $2\times T\times H\times W$.
In practice, we estimate a set of optical flows as the DF\textemdash$\mathbf{D}$ by E-RAFT~~\cite{gehrig2021raft}, as it has shown promising results in handling challenging motion scenes.
In this way, we can fit the nonlinear motion with the linear motions of $T$ segments.
If we know the location of a pixel at time $t_0$ and there is a considerable time gap between $t_0$ and $t_n$, we can obtain the location of this point at time $t_n$ by adding up its displacement in DF over the time interval, as Eq.~\ref{eq:inter}.
{\begin{equation}
    \begin{split}
    tr(t_n) &= tr(t_0) + \sum_{i=1}^{n} \left(tr(t_i) - tr({t_{i-1}})\right) \\
    &= tr(t_0) + \sum_{i=1}^{n} p_{{i-1}\to i}  .
    \end{split}
    \label{eq:inter}
\end{equation}}
\noindent\textbf{Displacement field Loss:} To encourage generating a smooth displacement field, we follow previous works~\cite{he2022timereplayer,jiang2018super} to promote the consistency in flow values between adjacent pixels, as in Eq.~\ref{eq:loss_df}, where $\nabla$ is a directional gradient and $T$ is the temporal dimension.
{\begin{equation}
    \mathbf{L}_{field} =
        \frac{1}{T} \sum_{i=1}^{T}\left((\nabla_{x}\mathbf{D}[i])^{2}+(\nabla_{y}\mathbf{D}[i])^{2}\right)
    \label{eq:loss_df}
\end{equation}
}

\begin{figure}[t!]
\centering
\includegraphics[width=\linewidth]{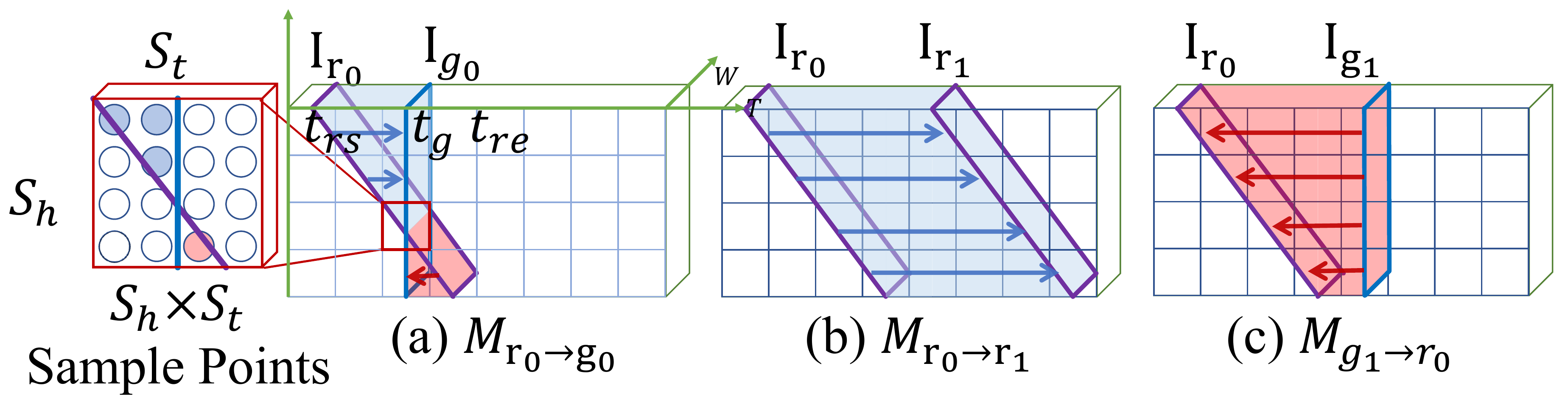}
\caption{{
An illustration of weight maps of RS-to-GS and RS-to-RS for the displacement field.
\textcolor{blue}{Blue} represents a positive value, indicating forward movement.
\textcolor{red}{Red} represents a negative value, indicating backward movement.
Each item in the weight map uses uniform sample $S_h\times S_t$ points to estimate its weight.
{Two time points define the rolling shutter exposure: the start time $t_{rs}$ and the end time $t_{re}$. In contrast, the global shutter frame is exposed at a single time point $t_g$.
Subfigures (a), (b), and (c) provide detailed examples: (a) shows the transformation from an RS frame to a GS frame, (b) shows an RS-to-RS mapping, and (c) illustrates the mapping from an RS frame to a preceding GS frame.}
}}
\label{fig:relation_rs_to_gs_mask}
\label{fig:mask}
\end{figure}

\begin{figure*}[t!]
\centering
\includegraphics[width=0.95\linewidth]{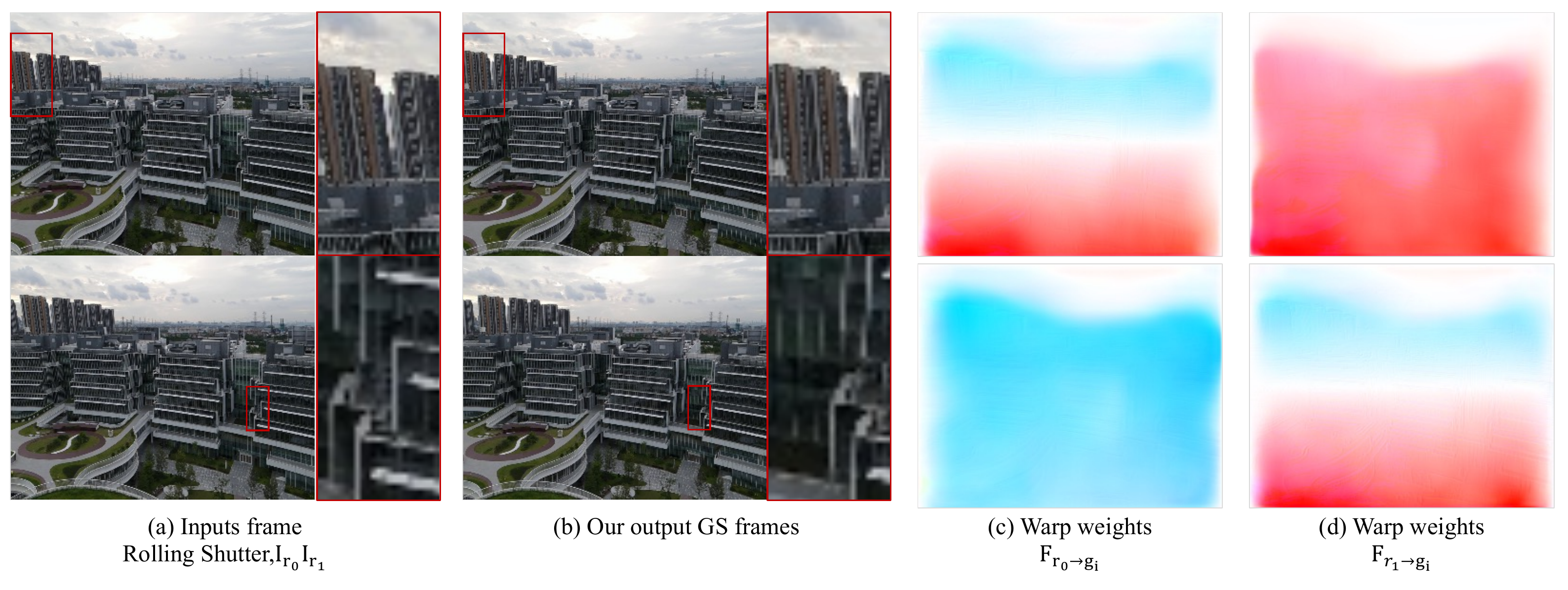}
\caption{{
Evaluation of our framework on the UAV-RS dataset, we utilize Fastec simulation dataset~\cite{Liu2020DeepSU} for training and test on this dataset.}
Displayed herein are the outcomes of both rolling shutter correction and frame interpolation.
Additionally, a visualization of the warp weight is presented.
From the visualization, it's evident that our framework yields generalization performance on the UAV dataset with deformation correction and frame interpolation.
For more vivid depict of  this scene, please visit the {\textbf{accompanying video in the supplementary material}}.}
\label{fig:UAV-rolling-currection}
\end{figure*}

\noindent\textbf{Weight map Design for RS GS transformation}
\label{sec:mask_design}
We designed the weight map based on the exposure mechanism of the RS and GS to describe the transition between any two frames.
The exposure model of the rolling shutter frame and events are shown in Fig.\ref{fig:relation_rs_to_gs_mask}.
The red solid line represents the start time of the frame exposure, and the red dashed line represents the end time of the frame exposure.
It can be easily found that the exposure time is much shorter than the rolling shutter time.
Long exposure time can lead to blurring, which complicates the research question of frame interpolation, and we do not consider the long exposure in the paper.
Therefore, we approximate the rolling shutter frame and the global shutter frame as planes.
Obviously, the plane of the global shutter can be described as $t=t_g$, where $t_g$ is the global shutter exposure timestamp.
The plane of the rolling shutter is parallel to the W-axis, {row of the image}, and passes through two points $(h, t_{rs})$ and $(0, t_{re})$ at the same time, where $h$ is the height of frames, and $t_{rs}$ and $t_{re}$ are begin time and end time of rolling shutter frame exposure.
Given two planes, we use uniform sampling to calculate the weights of each bin in each weight map.
For each sampling point, we calculate whether it is on the left or right of the plane by computational geometry.
Fig.~\ref{fig:mask} shows the projection of more weight maps in the $H\times T$ dimension.
Fig.~\ref{fig:UAV-rolling-currection} shows the outputs of our framework and the visualization of warp weight in an aerial scene.

\begin{figure}[t!]
\centering
\includegraphics[width=1\linewidth]{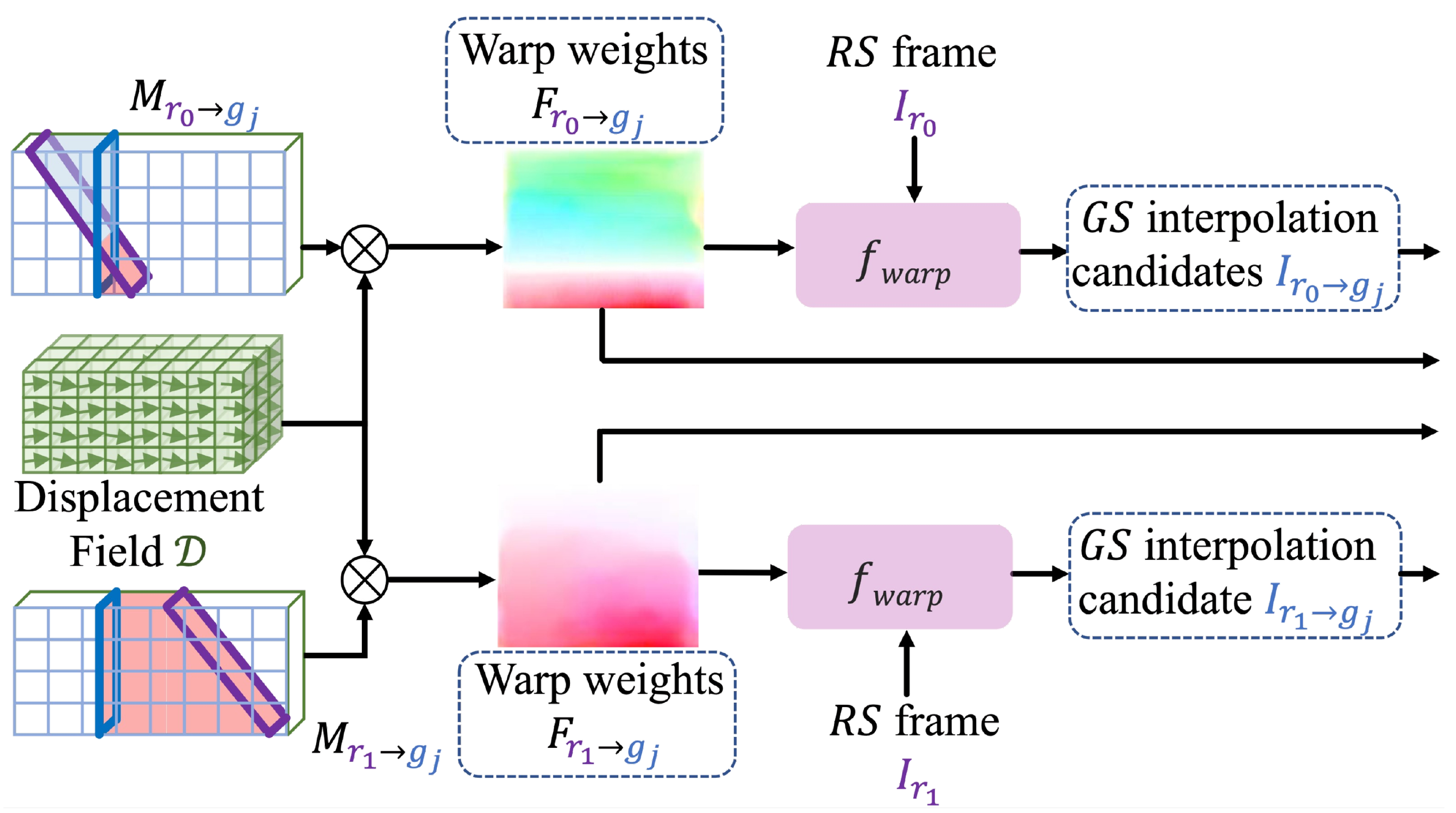}
\caption{The illustration of latent GS generation from RS frames $I_{r_0}$, $I_{r_1}$ and displacement field $\mathbf{D}$. The dash boxes indicate the output.}
\label{fig:rs0-gsi-Release}
\end{figure}

\subsection{Latent Global Shutter Frame Generation}
\label{sec:latent_gs_frame_generation}
This module aims to generate a series of latent GS frames based on RS frames and DF, as shown in Fig~\ref{fig:overall_framework} (II).
Based on the analysis in Sec.~\ref{sec:displacement_field_estimation}, it can be inferred that the value of any point in the video can be determined by the value of a known point and the corresponding transformation in the $\mathbf{D}$.
Therefore, weight maps are proposed to select the corresponding displacement field information for the reciprocal reconstruction between RS and GS frames, as shown in Fig.~\ref{fig:relation_rs_to_gs_mask}.
The shape of the weight map is $T\times H\times W$, and each item of the weight map is a weight of motion for the corresponding position in DF.
Specifically, we represent RS or GS frames as planes determined by their exposure time.
We sample uniformly inside each item in the weight map, and for each sampled point we calculate its direction to the RS and GS planes, as shown in Fig.\ref{fig:relation_rs_to_gs_mask}.
For example, Fig.~\ref{fig:relation_rs_to_gs_mask} (a) shows the weight map $\mathbf{M}_{{r_0}\to{g_0}}$ of RS frame $I_{r_0}$ to GS frame $I_{g_0}$.
Blue and red indicate the forward and backward deformation from $I_{r_0}$ to generate $I_{g_0}$.

{As illustrated in Fig.~\ref{fig:rs0-gsi-Release}, which depicts the schematic diagram of the entire process, we define the weight map that converts the $i$-th RS frame $I_{r_i}$ into the $j$-th GS frame $I_{g_j}$ as $\mathbf{M}_{{r_i}\to{g_j}}$.}
For the generation of GS frames, we use Eq.~\ref{eq:rs_to_gs_with_d_f} and Eq.~\ref{eq:rs_to_gs_with_d}, where $f_{\text{warp}}$ is the warping function~\cite{glasbey1998review}. $F_{{r_0}\to{g_j}}$ and $F_{{r_1}\to{g_j}}$ are the estimated flow fields used to warp the RS frames $I_{r_0}$ and $I_{r_1}$ toward the target GS frame $I_{g_j}$, and $I_{{r_i}\to{g_j}}$ denotes the reconstructed GS frame from the $i$-th RS frame.
{The warp fields $F_{{r_i}\to{g_j}}$ are obtained by combining the predicted displacement field $\mathbf{D}$ and a learned temporal weight map $\mathbf{M}$ through a weighted summation along the temporal axis. The operation is defined as Eq.~\ref{eq:rs_to_gs_with_d_f}.}
\begin{equation}
    \begin{split}
    F_{{r_0}\to{g_j}} = \mathbf{D} \otimes \mathbf{M}_{{r_0}\to{g_j}}, \quad
    F_{{r_1}\to{g_j}} = \mathbf{D} \otimes \mathbf{M}_{{r_1}\to{g_j}},
    \end{split}
    \label{eq:rs_to_gs_with_d_f}
\end{equation}

{
Here, $\mathbf{D} \in R^{2 \times T \times H \times W}$ is the displacement field, representing motion vectors in both horizontal and vertical directions over $T$ temporal bins.
The weight maps $\mathbf{M}_{{r_i}\to{g_j}} \in R^{T \times H \times W}$ assign importance to each displacement sample.
The operation $\otimes$ performs element-wise multiplication between the two tensors followed by summation along the temporal dimension $T$, yielding a final flow field $F \in R^{2 \times H \times W}$.}
Finally, each warped GS frame is obtained using the warping function as Eq.~\ref{eq:rs_to_gs_with_d}. This process enables temporally and geometrically consistent synthesis of global shutter images from temporally misaligned rolling shutter inputs.
\begin{equation}
    \begin{split}
    I_{{r_i}\to{g_j}} = f_{\text{warp}}(I_{r_i}, F_{{r_i}\to{g_j}}), \quad i \in \{0, 1\},
    \end{split}
    \label{eq:rs_to_gs_with_d}
\end{equation}

This way, we can predict an arbitrary number of GS frames from two input RS frames, respectively, as shown in Fig.~\ref{fig:overall_framework}(II).
However, due to the occlusions and viewing angles, the quality of the GS frame mapped from a single RS frame is not good enough; therefore, we propose a refine network $f_{refine}$ to fuse multiple GS frames.
For convenience, we employ U-Net, used by CVR~\cite{fan2022context}, as the refined network.
Specifically, we concatenate the $I_{r_0}$,$I_{r_1}$,$I_{{r_0}\to{gi}}$,$I_{{r_1}\to{g_j}}$,$F_{{r_1}\to{g_j}}$, and $F_{{r_0}\to{g_j}}$ in the channel dimension, so that the network can obtain RS frames and motion information from different perspectives, and fusion to reduce the occlusion effect.
The input and output of $f_{refine}$ are formulated in Eq.~\ref{eq:refine},
\begin{equation}
    \begin{split}
    & \Delta F_{{r_0}\to{g_j}}, \Delta F_{{r_1}\to {g_j}}, O_{g_j} = \\
    & f_{refine}(I_{{r_0}\to{g_j}}, I_{{r_1}\to{g_j}}, I_{r_0}, I_{r_1}, F_{{r_0}\to {g_j}}, F_{{r_1}\to{g_j}}),
    \end{split}
    \label{eq:refine}
\end{equation}
where $\Delta F_{{r_0}\to{g_j}}, \Delta F_{{r_1}\to{g_j}}$ are residuals of the warp weights $F_{{r_0}\to{g_j}}$ and $F_{{r_1}\to{g_j}}$, and $O_{g_j} \in [0,1]$ is the degree of occlusion for $I_{r_0}$. Therefore, the refined latent $j$-th GS frame is generated, as Eq.~\ref{eq:gs_fusion}:
{\begin{equation}
    \begin{split}
        I_{g_j} = &O_{g_j} \times f_{warp}(I_{r_0},F_{{r_0}\to{g_j}} + \Delta F_{{r_0}\to{g_j}}) + \\
    &(1 - O_{g_j}) \times f_{warp}(I_{r_1},F_{{r_1}\to{g_j}} + \Delta F_{{r_1}\to{g_j}}).
    \end{split}
    \label{eq:gs_fusion}
\end{equation}}

\subsection{Reciprocal Reconstruction}
\label{sec:self-supervision}
As the ground truth GS frames are not available and the mapping from the input RS frames to latent GS frames is highly under-constrained, this module exploits how to reconstruct RS frames from a single GS frame for the purpose of self-supervision.
As the displacement field includes the non-linear dense 3D spatiotemporal information of all pixels during the exposure time, we can simply achieve this target by Eq.~\ref{eq:gs_to_rs_warp_weight} and Eq.~\ref{eq:gs_to_rs}, where $\mathbf{M}_{{g_j}\to{r_i}}$ is the weight map from $j$-th GS frame $I_{g_j}$ to $i$-th RS frame $I_{r_i}$.
Based on the definition of weight map in Sec.~\ref{sec:displacement_field_estimation}, it indicates that weight maps of RS-to-GS and GS-to-RS are reversible for each other, namely $\mathbf{M}_{{g_j}\to{r_i}}=-\mathbf{M}_{{r_i}\to{g_j}}$.
Because we can reconstruct RS frames $I_{r_0}$ or $I_{r_1}$ from each predicted latent GS frame $I_{g_j}$, we convert the supervision from the reconstruction of latent GS frames to the reconstruction of input RS frame.
{\begin{equation}
    \begin{split}
    F_{{g_j}\to{r_0}} &= \mathbf{D} \otimes \mathbf{M}_{{g_j}\to{r_0}}, \\
    F_{{g_j}\to{r_1}} &= \mathbf{D} \otimes \mathbf{M}_{{g_j}\to{r_1}}
    \end{split}
    \label{eq:gs_to_rs_warp_weight}
\end{equation}}
{\begin{equation}
    I_{{g_j}\to{r_i}} = f_{warp}(I_{g_j}, F_{{g_j}\to{r_i}}), i=\{0,1\}
    \label{eq:gs_to_rs}
\end{equation}}
\noindent\textbf{(A) GS-to-RS Loss}:
To realize the self-supervision, we reconstruct the input RS frames $I_{r_0}$, $I_{r_1}$ from the generated $i$-th frame $I_{g_i}$. For simplicity, we use the {Charbonnier loss} $\mathbf{L}_{c}$ ~\cite{lai2018fast} as the GS-to-RS self-supervision loss, formulated as:
{\begin{equation}
    \mathbf{L}_{gs2rs} = \frac{1}{2N} \sum_{i=1}^{N} \left(\mathbf{L}_{c}(I_{{g_i}\to{r_0}},I_{r_0}) + \mathbf{L}_{c}(I_{{g_i}\to{r_1}},I_{r_1})\right),
    \label{eq:loss_gs2rs}
\end{equation}}
where $N$ is the number of predicted GS frames.

\begin{figure*}[t!]
\centering
\includegraphics[width=0.92\linewidth]{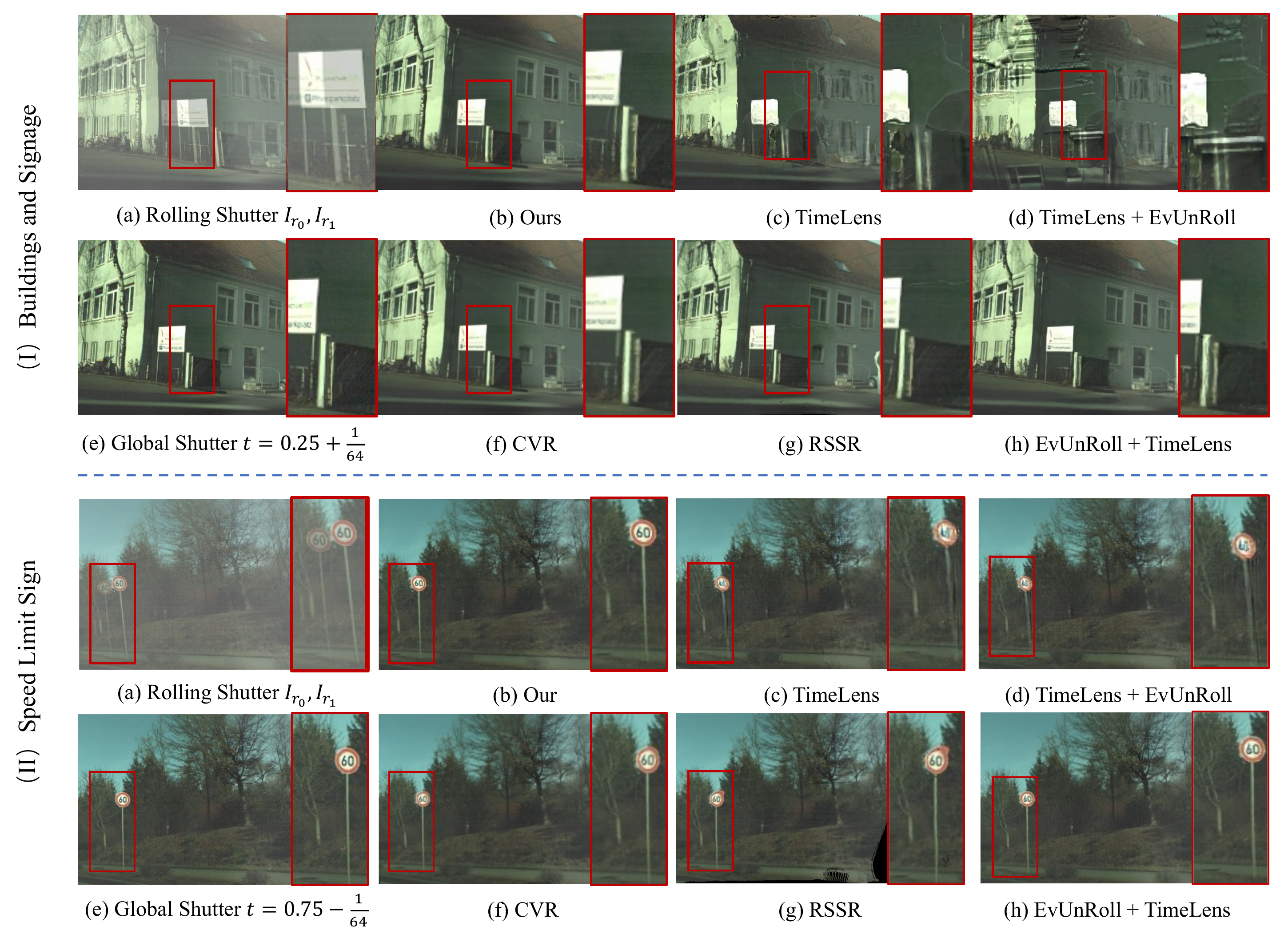}
\caption{
Visual comparison on Fastec-RS dataset~\cite{Liu2020DeepSU}, where two random time points are selected as target frames for recovery.
{The results of RSSR (g) exhibit black holes, particularly evident in scene (II). While TimeLens (c) and TimeLens+EvUnRoll (d) show noticeable distortion and artifacts when rolling shutter distortion is not accounted for. Our results demonstrate competitiveness with supervised methods (e.g., CVR and EvUnRoll + TimeLens), and even outperform them in accurately reconstructing the speed limit sign within scene (II).}
}
\label{fig:fastec-vis}
\end{figure*}

\begin{table*}[t!]
\centering
\caption{Quantitative results (PSNR~/~SSIM~/~LPIPS) of the proposed framework and other methods on the Fastec-RS\cite{Liu2020DeepSU} dataset. Bold indicates the best performance.}
\resizebox{\linewidth}{!}{
\begin{tabular}{r|r|cc|ccccc}
\toprule
Methods & Params(M)                                                    & Event             & SSL                         & 4$\times$              & 8$\times$ & 16$\times$   & 24$\times$            & 32$\times$ \\ \midrule \hline
CVR\cite{fan2022context} &42.69                                  & \ding{55}     & \ding{55}                 & 24.85~/~0.7538~/~0.1115    & 26.11~/~0.8003~/~0.1039      & 27.00~/~0.8330~/~0.0995   & 27.28~/~0.8434~/~0.0981       & 27.40~/~0.8481~/~0.0974        \\
RSSR~\cite{fan2021inverting} &26.04                                & \ding{55}     & \ding{55}                 & 18.61~/~0.5975~/~0.1808    & 18.41~/~0.5844~/~0.1858      & 18.32~/~0.5780~/~0.1888   & 18.28~/~0.5759~/~0.1899                          & 18.26~/~0.5747~/~0.1905         \\
TL~\cite{tulyakov2021time} &72.20                                 & \ding{51}     & \ding{55}                 &22.14~/~0.6334~/~0.1993                                                  &22.33~/~0.6408~/~0.1950 &22.34~/~0.6413~/~0.1938                      &22.38~/~0.6425~/~0.1933                           & 22.40~/~0.6432~/~0.1929      \\
TL~\cite{tulyakov2021time}+EU~\cite{zhou2022evunroll} &93.03       & \ding{51}     & \ding{55}                 &23.79~/~0.6739~/~0.1945                       &22.76~/~0.6376~/~0.2267                         &22.24~/~0.6182~/~0.2407                       &22.13~/~0.6141~/~0.2442                           & 22.08~/~0.6120~/~0.2463          \\
EU~\cite{zhou2022evunroll}+TL~\cite{tulyakov2021time}  &93.03      & \ding{51}     & \ding{55}                 &\textbf{28.44}~/~\textbf{0.8450}~/~0.0991    &\textbf{28.67}~/~\textbf{0.8487}~/~0.0983      & \textbf{28.75}~/~\textbf{0.8504}~/~0.0980   & \textbf{28.77}~/~\textbf{0.8507}~/~0.0977       & \textbf{28.78}~/~\textbf{0.8510}~/~0.0977        \\ \hline
EU~\cite{zhou2022evunroll}+TR~\cite{he2022timereplayer} &80.38     & \ding{51}     & EU\ding{55},TR\ding{51}   & 21.55~/~0.6149~/~0.1624    & 21.94~/~0.6318~/~0.1559      & 22.21~/~0.6431~/~0.1529   & 22.31~/~0.6478~/~0.1518       & 22.36~/~0.6499~/~0.1510         \\
Our &22.00                                                         & \ding{51}     & \ding{51}                 & 26.27~/~0.8086~/~\textbf{0.0834}    & 26.29~/~0.8095~/~\textbf{0.0827}      & 26.26~/~0.8034~/~\textbf{0.0810}   &  26.37~/~0.8049~/~\textbf{0.0853}      & 26.31~/~0.8074~/~\textbf{0.0836}     \\ \bottomrule
\end{tabular}
}
\label{tab:FasCompare}
\end{table*}

\noindent\textbf{(B) RS-to-RS Loss}:
Since the two input RS frames have spatiotemporal coherence, we leverage it as the constraint for imposing additional self-supervision.
We reconstruct $i$-th RS frame by warping $j$-th RS frame and displacement field as Eq.~\ref{eq:rs1_to_rs0} and  Eq.~\ref{eq:rs0_to_rs1}.
Then, we employ the $\mathbf{L}_c$ as our RS-to-RS loss as Eq.~\ref{eq:rs2rs}.
{\begin{equation}
I_{{r_1}\to{r_0}}=f_{warp}\left(I_{r_1}, (\mathbf{D} \otimes \mathbf{M}_{{r_1}\to{r_0}})\right),
\label{eq:rs1_to_rs0}
\end{equation}}
{\begin{equation}
I_{{r_0}\to{r_1}}=f_{warp}\left(I_{r_0}, (\mathbf{D} \otimes \mathbf{M}_{{r_1}\to{rs_0}})\right),
\label{eq:rs0_to_rs1}
\end{equation}}
{\begin{equation}
    \mathbf{L}_{rs2rs} = \mathbf{L}_{c}(I_{{r_0}\to{r_1}},I_{r_1}) + \mathbf{L}_{c}(I_{{r_1}\to{r_0}},I_{r_0})
    \label{eq:rs2rs}
\end{equation}}
\noindent\textbf{Total Loss:}
Finally, the total loss can be summarized as Eq.~\ref{eq:loss_all}, where $\lambda_f$, $\lambda_{rs}$,$\lambda_{gs}$ denote the weights of each loss.
{\begin{equation}
    \mathbf{L} = \lambda_f \mathbf{L}_{field} + \lambda_{rs} \mathbf{L}_{rs2rs} + \lambda_{gs} \mathbf{L}_{gs2rs}
    \label{eq:loss_all}
\end{equation}}

\begin{figure*}[ht!]
\centering
\includegraphics[width=0.9\linewidth]{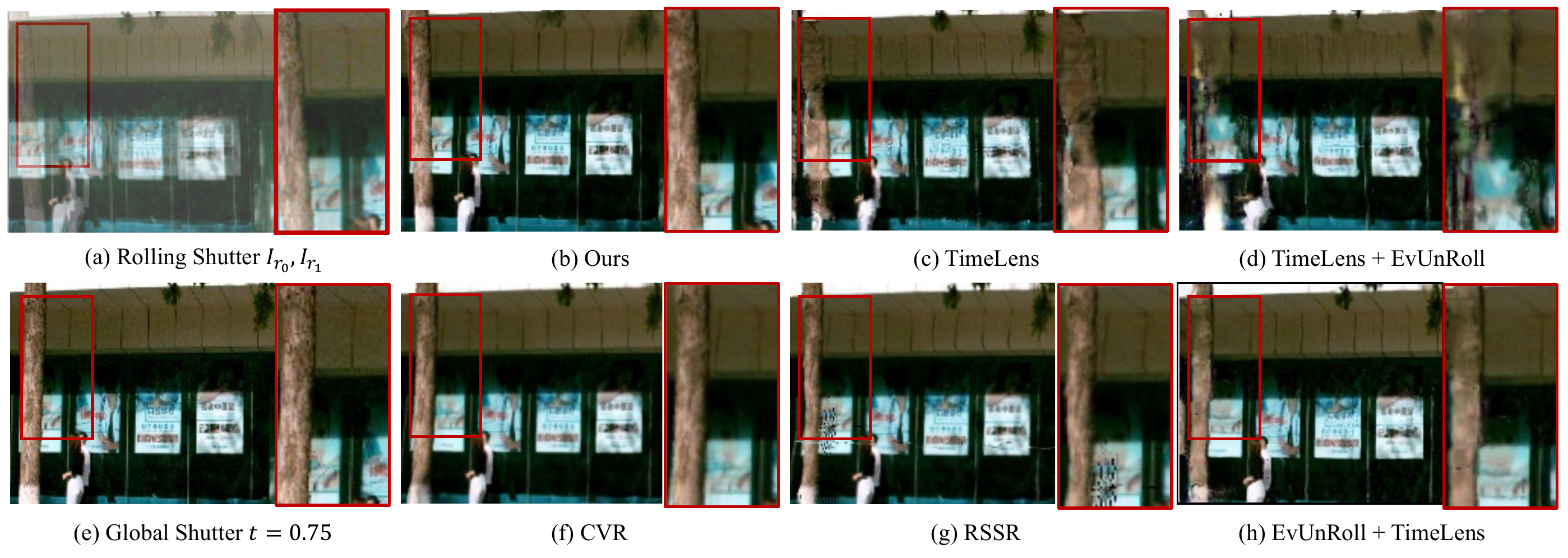}
\caption{Visual comparison on Gev-RS dataset~\cite{zhou2022evunroll}.
{{Similar to the results on the Fastec-RS dataset,} RSSR's results (g) exhibit black holes, whereas TimeLens (c) and TimeLens+EvUnRoll (d) show noticeable distortion and artifacts. Importantly, our results accurately reconstruct the details of the poster (highlighted in the red box area), especially the face that other supervised methods find challenging to accomplish.}
}
\label{fig:esp-gev}
\end{figure*}

\begin{table*}[ht]
\centering
\caption{
Quantitative results (PSNR~/~SSIM~/~LPIPS) of the proposed framework and other methods on the Gev-RS simulated dataset.
For supervised methods, TL refers to TimeLens\cite{tulyakov2021time}, EU refers to EvUnRoll\cite{zhou2022evunroll}.
For the unsupervised method, TR refers to TimeReplayer\cite{he2022timereplayer}.
LPIPS is calculated by the \href{https:~/~~/~github.com~/~richzhang~/~PerceptualSimilarity}{lpipis} library\cite{zhang2018perceptual} with AlexNet pretrained model\cite{krizhevsky2017imagenet}.
'Event' indicates whether events are used, and
'SSL' indicates whether it is self-supervised. Black bold indicates the best performance.
}
\renewcommand{\tabcolsep}{3pt}
\resizebox{\linewidth}{!}{
\begin{tabular}{r|r|cc|ccccc}
 \toprule
Methods                                             & Params(M)    & Event             & SSL                         & $4\times$                        & $8\times$                 & $16\times$           & $24\times$            & $32\times$ \\  \midrule \hline
CVR\cite{fan2022context}                            & 42.69   & \ding{55}     & \ding{55}                 & 22.59~/~0.7508~/~0.1094                     & 23.80~/~0.7949~/~0.1027       & 24.60~/~0.8209~/~0.0990   & 24.85~/~0.8291~/~0.0979   & 24.98~/~0.8331~/~0.0973\\
RSSR\cite{fan2021inverting}                         & 26.04   & \ding{55}     & \ding{55}                 & 17.83~/~0.5875~/~0.1498                     & 17.58~/~0.5762~/~0.1532       & 17.45~/~0.5701~/~0.1553   & 17.40~/~0.5680~/~0.1560   & 17.37~/~0.5670~/~0.1562              \\
TL\cite{tulyakov2021time}                           & 72.20   & \ding{51}     & \ding{55}                 & 19.77~/~0.6408~/~0.1563                     & 19.86~/~0.6476~/~0.1518        &19.88~/~0.6492~/~0.1506    & 19.90~/~0.6128~/~0.1691   & 20.01~/~0.6526~/~0.1492  \\
TL\cite{tulyakov2021time}+EU\cite{zhou2022evunroll}  & 93.03   & \ding{51}     & \ding{55}                & 21.09~/~0.6682~/~0.1696                     & 20.05~/~0.6267~/~0.1981        &19.57~/~0.6079~/~0.2101    & 19.44~/~0.6027~/~0.2134   & 19.39~/~0.6002~/~0.2152    \\
EU\cite{zhou2022evunroll}+TL\cite{tulyakov2021time}  & 93.03  & \ding{51}     & \ding{55}                 & \textbf{25.62}~/~\textbf{0.8339}~/~0.0716   & \textbf{25.28}~/~\textbf{0.8290}~/~\textbf{0.0716}       & \textbf{25.30}~/~\textbf{0.8300}~/~0.0712   & \textbf{25.31}~/~\textbf{0.8306}~/~0.0709   & \textbf{25.32}~/~\textbf{0.8306}~/~0.0709     \\
\hline
EU\cite{zhou2022evunroll}+TR\cite{he2022timereplayer} & 80.38  & \ding{51}     & EU\ding{55},TR\ding{51}  & 19.02~/~0.6005~/~0.1612                     & 19.41~/~0.6210~/~0.1564       & 19.64~/~0.6321~/~0.1536   & 19.72~/~0.6362~/~ 0.1527  & 19.77~/~0.6382~/~0.1522    \\
Our                                                  & 22.00  & \ding{51}     & \ding{51}                 & 23.91~/~0.8091~/~\textbf{0.0662}            & 23.60~/~0.7973~/~0.0726       & 23.64~/~0.7964~/~\textbf{0.0702}   & 23.75~/~0.7971~/~\textbf{0.0699}   & 23.88~/~0.8074~/~\textbf{0.0702}     \\
\bottomrule
\end{tabular}
}
\label{tab:GevCompare}
\end{table*}

\section{Experiments}
\noindent\textbf{Implementation Details:}
We employ the Adam optimizer~\cite{kingma2014adam} for all experiments, with learning rates of $1e-4$ for all datasets. Our framework is trained for 100 epochs with a batch size of 4 using an NVIDIA RTX A30 GPU.

\noindent\textbf{Evaluation Metrics:} We evaluate our approach using the peak-signal-to-noise ratio (PSNR)~\cite{hore2010image} and structural similarity (SSIM)~\cite{wang2004image} and perceptual similarity metric LPIPS~\cite{zhang2018perceptual}.
\noindent\textbf{Dataset:}
\noindent\textbf{1) Fastec-RS dataset~}
Fastec-RS dataset~\cite{Liu2020DeepSU} provides the original frame sequences which are recorded by the high-speed GS cameras with the resolution of $640\times480$ at 2400 fps.
This dataset offers an external perspective of a fast-moving vehicle and can partially capture scenes resembling those of high-speed flying aircraft.
We first downsample the original videos to the same resolution($260\times346$) of the DAVIS346 event camera~\cite{scheerlinck2019ced}.
Then, we input the resized frames to the event simulator vid2e~\cite{gehrig2020video} to synthesize event streams.
We generate RS frames based on the same RS simulation process of Fastec-RS~\cite{Liu2020DeepSU}.
Besides, we employ the same dataset split scheme as in Fastec-RS~\cite{Liu2020DeepSU}: 56 sequences for training and 20 sequences for testing.
\textbf{2) Gev-RS dataset~}\cite{zhou2022evunroll}
provides original videos captured by GS high-speed cameras with $1280\times720$ resolution at 5700 fps.
We follow the aforementioned settings to downsample frame sequences and generate the corresponding event streams and RS frames.
We follow the same dataset split scheme as in EvUnroll~\cite{zhou2022evunroll}: 20 videos for training and nine videos for testing.
Notably, EvUnroll~\cite{zhou2022evunroll} considers both RS correction and deblurring. Thus, their simulation dataset includes blurry frames and lacks high frame rate frames for VFI evaluation.
Therefore, we reconstruct RS frames and events from original videos to avoid the influence of blurring.
\noindent\textbf{3) ERS dataset}
To evaluate our method on the real-world dataset, we use an ALPIX-Eiger event camera~\cite{yunfan2024rgb} to collect a new dataset called ERS.
This camera outputs RGB frames with the resolution of $3264\times 2448$ and events with the resolution of $1632\times 1224$.
For all collected videos, 19 videos are selected for training, and ten videos are for testing. Finally,  3630 and 2071 frames with aligned events are used as the training and testing sets, respectively. To prevent memory overflow during training, we apply data augmentation strategies of \cite{fan2021inverting}, such as random crop, to our ERS dataset for all compared methods.
\noindent\textbf{4) UAV-RS dataset}
The first two datasets mentioned above have been widely adopted as benchmarks for quantitative evaluations in the academic community. In contrast, our third dataset is designed to facilitate real-world qualitative assessments. To comprehensively evaluate the robustness and generalizability of our method in various aerial photography scenarios, we introduced a new dataset based on aerial photography. We deployed a DJI drone to capture nine high-frame-rate videos with $120 fps$ frame rate, from which we generated simulated events and RS frames.
{The resolution of the generated rolling shutter frame is 260 x 346, with a frame rate of 0.46 fps (calculated as 120/260).}
This allowed us to conduct a qualitative evaluation of our method's effectiveness in aerial imaging contexts.

\subsection{Comparison with SoTA Methods}
{We compare our method with five SoTA methods under three VFI settings:
one SoTA event-guided VFI method\textemdash TimeLens~(TL)~\cite{tulyakov2021time}, which is based on GS frames.
Combined methods, event-guided RS correction method\textemdash EvUnRoll (EU)~\cite{zhou2022evunroll} + event-guided VFI methods: TL \cite{tulyakov2021time} or TimeReplayer (TR)~\cite{he2022timereplayer}.
Frame-based RS-to-GS VFI methods: CVR~\cite{fan2022context} and RSSR~\cite{fan2021inverting}.
Except for the frame-based methods which only takes RS frames, the inputs of other methods are both RS frames and events, and the ground truth frames of these three settings are the GS frames.}

\noindent\textbf{Evaluation on Fastec-RS dataset:}
We evaluate our methods on the Fastec-RS dataset, and the quantitative result is summarized in Tab.~\ref{tab:FasCompare}. We draw a similar conclusion as the experiments on the
Gev-RS dataset, based on the quantitative comparison.
Fig.~\ref{fig:fastec-vis} shows the visualization results of an outside street view captured by a moving camera and we have the sharpest reconstruction, for example, the number on the road sign (in the red box).
{For more visualization results, please refer to the supplementary material (Faster.zip).}

\begin{figure*}[t!]
\centering
\includegraphics[width=0.9\textwidth]{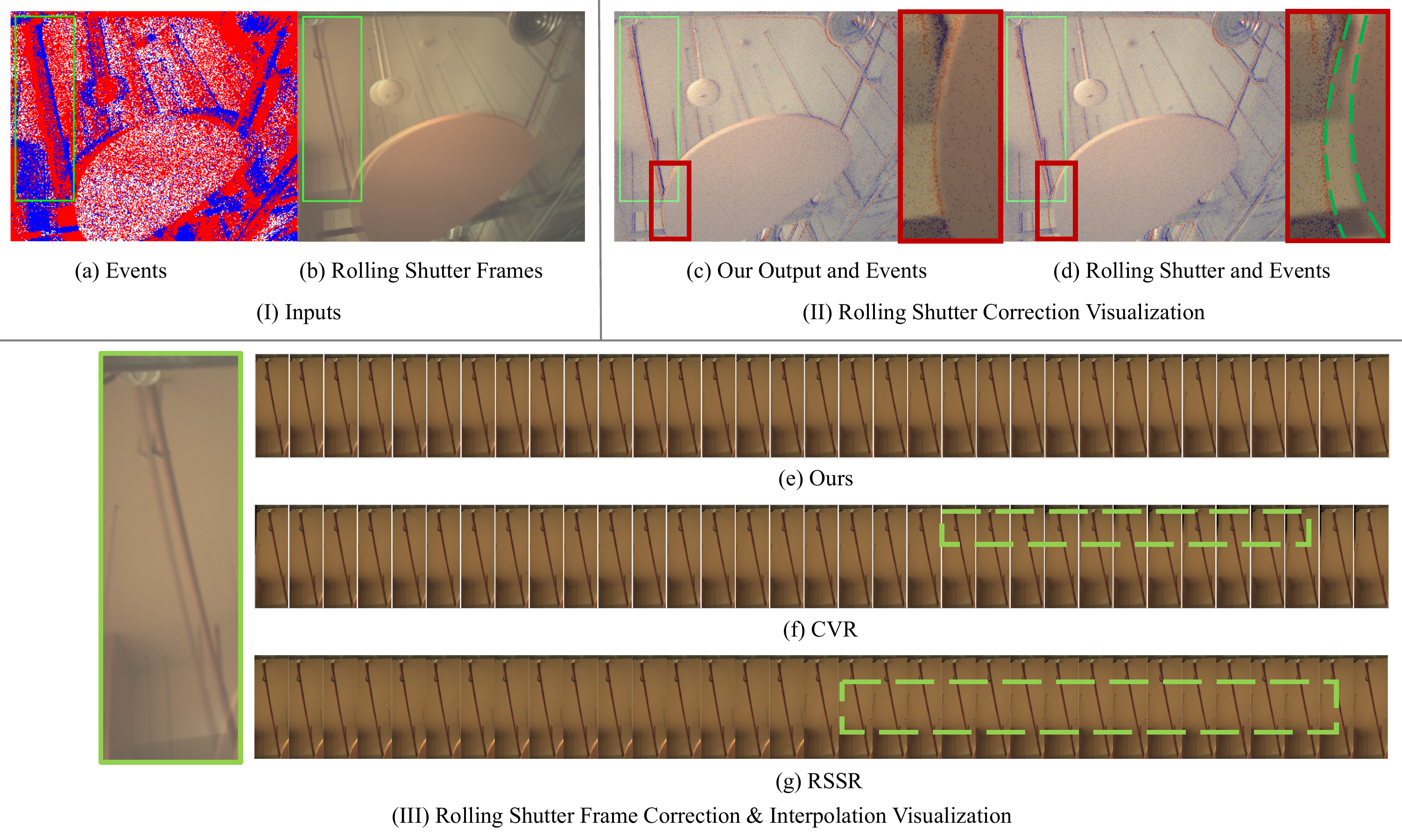}
\caption{{Visualization results from real data: (I) showcases the input, consisting of events and two successive RS frames. Following RS correction, the output is presented in (II). By overlaying the events onto both our output and the initial input, it becomes evident that our method adeptly rectifies RS deformations. The interpolated frame output is depicted in (III), where our technique achieves a 32-fold interpolation.{The green dashed boxes highlight temporal artifacts that appear in competing methods. Please zoom in to observe these details more clearly.}}}
\label{fig:14-aplex-visualization-release}
\end{figure*}

\noindent\textbf{Evaluation on ERS dataset}
By comparing the edge in the RS frames with that of events, we successfully correct the distorted edges, as shown in Fig.\ref{fig:14-aplex-visualization-release} (a).
{For more visualization results, please refer to the supplementary material (DJI.zip).}

\noindent\textbf{Evaluation on Gev-RS dataset:}
Tab.~\ref{tab:GevCompare} presents the quantitative results from $4\times$ to $32\times$ interpolation, and the comparison of visual quality is shown in Fig.~\ref{fig:esp-gev}.
Our method clearly outperforms the CVR and RSSR, which are frame-based RS-to-GS supervised VFI methods, in $4\times$ interpolation by up to \textbf{1.32 dB} in PSNR.
In addition, our method has the best LPIPS scores among all the compared methods (expect $8\times$ interpolation).
Because our results do not suffer from the black holes, as shown in the red box of Fig.~\ref{fig:esp-gev}~(g).
This could be attributed to the capacity to estimate the complex non-linear motion of the displacement field by utilizing the high temporal resolution of events.
{For more visualization results, please refer to the supplementary material (Gev-RSC.mp4).}

\subsection{Efficiency Evaluation}

In this section, we delve into an evaluation of our method, focusing primarily on its generalization, inference speed, transmission bandwidth, {etc}.
This analysis aims to provide a comprehensive understanding of the efficiency of our model and its robustness.

\noindent\textbf{Generalization testing on UAV-RS dataset:}
{In Fig.~\ref{fig:UAV-rolling-currection}, we present the visualization of our model on the UAV-RS dataset. }
Having been trained on the Fastec dataset, the model's evaluation on the UAV dataset serves to underscore its generalization capabilities. Impressively, our approach adeptly rectifies the RS deformation, and the outcomes of the frame interpolation are notably effective.

\begin{table}[t!]
\caption{Ablation results on Gev-RS. $T$ indicates the count of time bins. $P$ indicates the use of pre-train optical flow model. $\mathbf{L}_p$ indicates the perceptual loss. $S_h$ and $S_t$ indicate the sample points for weight map.}
\resizebox{\linewidth}{!}{
\begin{tabular}{l|ccccccccc}
\toprule
& $T$ & P & \makecell[c]{$\mathbf{L}_p$ \\GS-to-RS} & \makecell[c]{$\mathbf{L}_p$ \\RS-to-RS} & $S_h$  & $S_t$  & PSNR  & SSIM  & LPIPS \\
\hline
\hline
1& 6        & \ding{51} & \ding{55}                        & \ding{55}                        & 50  & 100 & 23.91 & 0.8091 & 0.0662 \\
2& {\color{purple}\textbf{4}}        & \ding{51}        & \ding{55}                        & \ding{55}                        & 50  & 100 & 23.78 & 0.8019 & 0.0669\\
3& {\color{purple}\textbf{2}}        & \ding{51}        & \ding{55}                        & \ding{55}                        & 50  & 100 & 23.32 & 0.7928 & 0.0718 \\
4& {\color{purple}\textbf{12}}       & \ding{51}        & \ding{55}                        & \ding{55}                        & 50  & 100 & 23.67 & 0.7960 & 0.0674\\

\hline
5& 6        & {\color{purple}\textbf{\ding{55}}}        & \ding{55}                        & \ding{55}                        & 50  & 100 & 23.88 & 0.8082 & 0.0708\\
\hline
6& 6        & \ding{51}        & {\color{purple}\textbf{\ding{51}}}                        & \ding{55}                        & 50  & 100 & 18.50 & 0.6225 & 0.1791 \\
7& 6        & \ding{51}        & {\color{purple}\textbf{\ding{51}}}                        & {\color{purple}\textbf{ \ding{51}}}                        & 50  & 100 & 23.84 & 0.8134 & 0.0663 \\  \hline
8& 6        & \ding{51}        & \ding{55}                        & \ding{55}              & {\color{purple}\textbf{5}}   & {\color{purple}\textbf{10}}  & 23.90 & 0.8097 & 0.0666 \\
9& 6        & \ding{51}        & \ding{55}                        & \ding{55}              & {\color{purple}\textbf{100}} & {\color{purple}\textbf{200}} & 24.00 & 0.8109 & 0.0629 \\
\bottomrule
\end{tabular}
}
\label{tab:ablation_studies}
\end{table}

\begin{figure}[t!]
\centering
\includegraphics[width=0.9\linewidth]{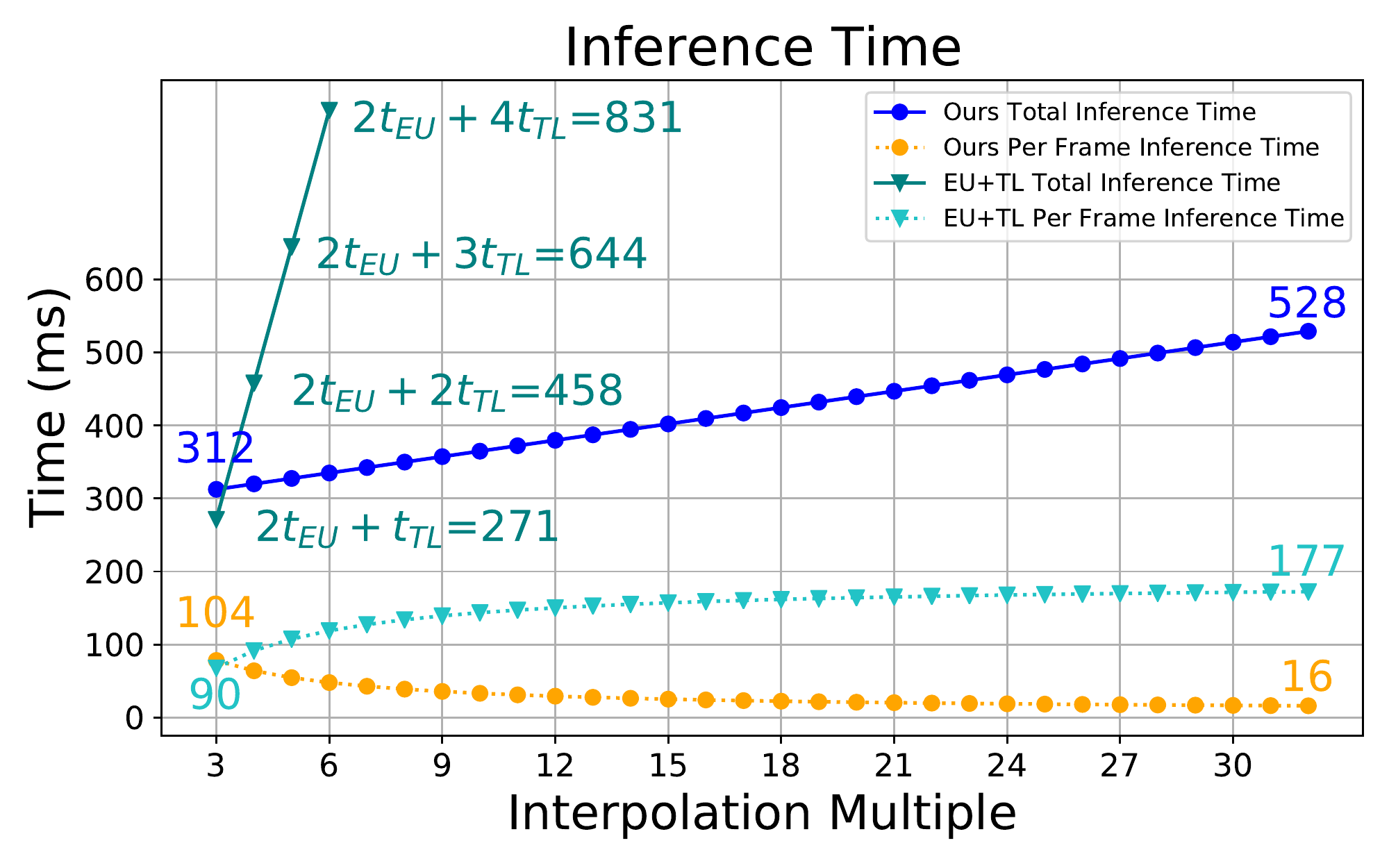}
\caption{
Comparison of the inference time of our method with TimeLens + EvUnRoll. We only count the time computing in the GPU, and I/O time is not included here. The x-axis is the magnification of the interpolated frame, from 3 times to 32 times. The y-axis represents time in milliseconds.}
\label{fig:interpolation-times-cost}
\end{figure}

\begin{figure*}[t!]
\centering
\includegraphics[width=0.92\linewidth]{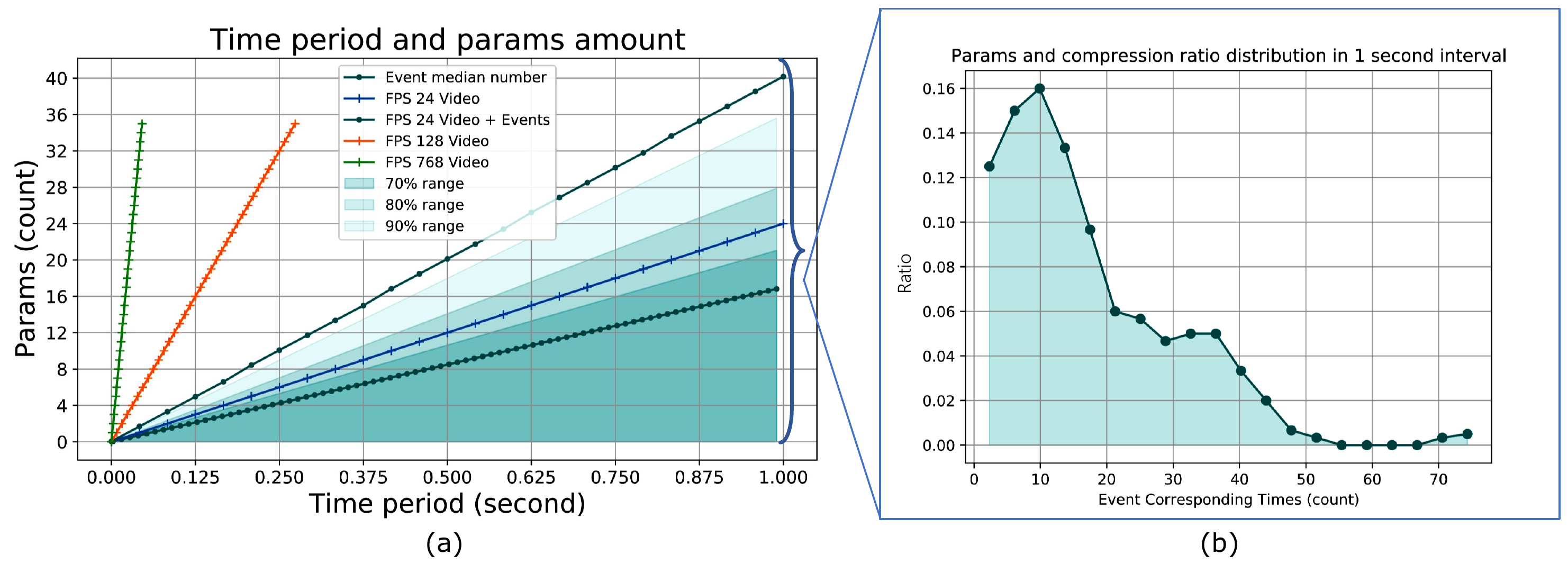}
\caption{{
(a) The number of parameters required over time for event data and videos at different frame rates (fps). Parameters refer to the raw data size, such as for an $N$-frame video with a resolution of $H\times W$, the parameters are calculated as $N\times H×W$. For events, the parameters are the number of event signals multiplied by 4. The x-axis represents the recording time from 0 to 1 second, and the y-axis shows the amount of parameters required to store the data. The red and blue lines represent videos recorded at 128 fps and 24 fps, respectively, with a linear increase over time. The black line represents the average number of events, where the variability of the event recording depends on the scene. The different cyan-shaded areas show the range of this variation.
(b) The distribution of event trigger times within 1 second.
Specifically, in one second, the ratio of a pixel outputting 10 event signals is 0.16, the ratio of outputting 30 signals is about 0.05, and only less than 0.01 pixels output more than 70 signals.
}}
\label{fig:event-amount-rate}
\end{figure*}

\noindent\textbf{Inference speed:}
To evaluate the efficiency of our method, we conducted the experiment on $32\times$ interpolation with the resolution of $260\times346$, as depicted in Fig.~\ref{fig:interpolation-times-cost}.
Evidently, with the increment in interpolation factors, the inference time for EU+TL exhibits a substantial increase, whereas our method demonstrates only a marginal escalation from \textbf{312ms} to \textbf{528ms}.
Notably, for a $32\times$ interpolation, our technique requires merely \textbf{16ms} to reconstruct each GS frame, which is approximately \textbf{one-tenth} of the time taken by EU+TL.

\noindent\textbf{Transmission bandwidth analysis:}
We analyzed the correlation between the number of event activations and time using real data, and the results are visualized in the accompanying Fig.~\ref{fig:event-amount-rate}.
In subfigure (a), the x-axis denotes time, while the y-axis signifies the average number of parameters needed per pixel. Notably, each activation requires one parameter for recording. The data suggests a linear increase in the number of activations over time. The median frequency of activations is approximately 17 times per second. Furthermore, 90\% of the pixels activate fewer than 36 times, 80\% activate less than 28 times, and 70\% activate fewer than 24 times.
Fig.~\ref{fig:event-amount-rate}~(b) illustrates the distribution of pixel activations within a 1-second duration. Notably, pixels activated 10 times constitute the majority at 16\%, whereas those activated 50 times represent less than 10\%.

By comparing the per-pixel params of event+fps 24 videos and fps 128 videos, we highlight the advantage of applying an event camera to reconstruct high-frame-rate video, especially with the consideration of limited transmission bandwidth. As illustrated in Fig.~\ref{fig:event-amount-rate}, the parameters of the event+fps 24 video are significantly smaller compared to those of the fps 128 video. This finding indicates that the event+fps 24 video is not only capable of providing sufficient spatial-temporal information to reconstruct high-frame-rate GS videos but also effectively mitigates the demand for transmission bandwidth.

{
\noindent \textbf{Model complexity:}
Compared with other methods, our model stands out for its advantage in parameters and algorithmic complexity.
Specifically, the parameter number of our framework is only one-fifth of EvUnRoll+TimeLens, as shown in Tab.~\ref{tab:GevCompare}.
In addition, during the inference stage, our approach only needs to estimate DF once for generating all GS frames.
In contrast, TimeLen and TimeReplayer require individual calculations of optical flow for each frame, causing high computation costs.
}

\begin{table}[t]
\caption{
Ablation of losses and supervised training for the 4 $\times$ interpolation on the Gev-RS dataset.}
\centering
\begin{tabular}{c|ccccccc}
\hline
& $\mathbf{L}_{gs2rs}$ & $\mathbf{L}_{field}$   & $\mathbf{L}_{rs2rs}$   & $\mathbf{L}_{gs}$ & PSNR      & SSIM   & LPIPS  \\
\hline
\hline
1 & \ding{51}           & \ding{55}               & \ding{55}               &\ding{55}           & 23.79	 & 0.8076 & 0.0672  \\
2 & \ding{51}           & \ding{51}               & \ding{55}               & \ding{55}          & 23.91	 & 0.8062 & 0.0680  \\
3 & \ding{51}           & \ding{51}               & \ding{51}               & \ding{55}          & \underline{23.91}     & \underline{0.8091} & \underline{0.0662} \\
4 & \ding{51}           & \ding{51}               & \ding{51}               & \ding{51}          & \textbf{25.08} & \textbf{0.8319} & \textbf{0.0647} \\
\hline
\end{tabular}
\label{tab:loss-ablation}
\end{table}

\subsection{Ablation Studies}
We conduct the following ablation experiments to evaluate the effectiveness of our proposed modules in Gev-RS dataset~\cite{zhou2022evunroll}:
E-RAFT Channel,
Time bins and iters of E-RAFT are set to 15, 6, and 12 in the baseline case which uses pretrained E-RAFT model~\cite{gehrig2021raft}. \\
\noindent\textbf{Loss function:}
Tab.\ref{tab:loss-ablation} for each loss in Eq.~14.
The 1st row represents the baseline, where solely the $\mathbf{L}_{gs2rs}$ loss (Eq.~10) is utilized, In the 2nd and 3rd rows, the displacement field loss (Eq.~3) and RS-to-RS loss (Eq.~13) are incorporated concurrently.
As a result, increases are observed in the PSNR (\textbf{+0.12}) and SSIM (\textbf{+0.0015}). Also, when perceptual loss is added in our self-supervised setting (Tab.~4), we find it hard to obtain favorable outcomes. \\
{
\noindent\textbf{Perceptual loss:}
Perceptual loss $\mathbf{L}_p$ is widely used in previous work\cite{he2022timereplayer,tulyakov2021time,jiang2018super,fan2021inverting,fan2022context}.
We also perform experiments to study the effectiveness of the perceptual loss~\cite{johnson2016perceptual} by applying the perpetual loss on GS-to-RS reconstruction and RS-to-RS warping.
Tab.~\ref{tab:ablation_studies} (row 6-7) validates that introducing the perceptual loss can not boost the performance.
Especially, applying perceptual loss only on the GS-to-RS reconstruction leads to the collapse of the displacement field and a worse performance.
We argue that although the perceptual loss is applied in the compared method, it can not serve as the constraint in our self-supervised method. } \\
\noindent\textbf{Time bins:}
We conducted the experiments to evaluate the influences of different time bins on the displacement field estimation.
As shown in Tab.~\ref{tab:ablation_studies} (row 1-4), the settings with six-time bins obtain the best PSNR and SSIM scores.
The explanation is that when the number of time bins goes up, the 3D deformation field will be divided into more segments bringing better performance because of more capacity of the non-linear motion.
However, an excessive number of time bins can result in accumulated errors of motion and the degradation of performance.
Fig.~\ref{fig:UAV-rolling-currection} shows warp weights corresponding to GS at different times.
It can be observed that the nonlinear motion of the bus is estimated clearly. \\
\noindent\textbf{Pretrained optical flow model:}
We evaluate the effect of the pre-train optical flow model by removing E-RAFT pre-train model.
As shown in Tab.~\ref{tab:ablation_studies} (row 1,5), the experiments with the pre-train model and without it demonstrate a similar performance. This indicates that our method can learn the optical flow without supervision. \\
\noindent\textbf{Sample points of height and time of weight map:}
We investigate the impact of different sample points in height and time dimension during the estimation of the weight map, as shown in Fig.~\ref{fig:relation_rs_to_gs_mask}. Tab.~\ref{tab:ablation_studies} (row 1,8-9) shows that the settings with different sample points have similar performance. This is because, for the setting with five height samples points and ten time samples points, each element has up to 50 samples points; thus, each element has ample samples to estimate the weight of warping.

\section{Conclusion}
In summary, we presented a novel method for reconstructing high-frame-rate (slow-motion) videos free from skew and jelly effects, thus enhancing the VR experience, which was previously constrained by consumer cameras hardware limitations.
Our approach enables arbitrary frame rate video frame interpolation (e.g., 32×) and reciprocal reconstruction between RS and GS frames through self-supervised learning.
\noindent\textbf{Limitation and Future Work}
As the first attempt to employ a self-supervised approach for VFI based on RS images and events, our study has yielded promising results on simulated data.
However, in the absence of quantitative metrics on real-world data, the efficacy of our method remains to be fully evaluated.
Further research will consider how to combine event cameras with consumer cameras, \eg, UAVs, to collect aligned real-world RGB frames and events.

\section*{Acknowledgments}
This work is supported by the National Natural Science Foundation of China (NSF) under Grant No. 62206069 (affiliated with Guangzhou HKUST
FYTRI), the MOE AcRF Tier 1 SSHR-TG Incubator Grant FY24 under Grant No. RSTG7/24, and the Open Project Program of State Key Laboratory of Virtual Reality Technology and Systems, Beihang University (No. VRLAB2025C04).

\bibliographystyle{IEEEtran}
\bibliography{egbib}

\end{document}